\documentclass[10pt,conference]{IEEEtran}
\def\IEEEtitletopspace{20pt}
\IEEEoverridecommandlockouts
\usepackage{cite}
\usepackage{amsmath,amssymb,amsfonts}
\usepackage{algorithmic}
\usepackage{graphicx}
\usepackage{textcomp}
\usepackage{xcolor}
\def\BibTeX{{\rm B\kern-.05em{\sc i\kern-.025em b}\kern-.08em
    T\kern-.1667em\lower.7ex\hbox{E}\kern-.125emX}}

\usepackage{pifont}
\usepackage{subcaption}
\usepackage{siunitx}
\usepackage{graphicx}
\usepackage[hyphens]{url}
\usepackage{hyperref}
\hypersetup{
    colorlinks=true,
    linkcolor=magenta,
    urlcolor=magenta
}
\usepackage{tabularx}
\usepackage{booktabs} 
\usepackage{multirow}
\usepackage{utfsym}
\usepackage{amsmath}
\usepackage{booktabs, caption, makecell}
\usepackage{threeparttable}
\newcolumntype{Y}{>{\centering\arraybackslash}X}

\begin{document}

\title{CoSense3D: an Agent-based Efficient Learning Framework for Collective Perception 
}

\makeatletter
\newcommand{\linebreakand}{%
  \end{@IEEEauthorhalign}
  \hfill\mbox{}\par
  \mbox{}\hfill\begin{@IEEEauthorhalign}
}
\makeatother

\author{
\and
\IEEEauthorblockN{Yunshuang Yuan}
\IEEEauthorblockA{
\textit{Institute of Cartography and Geoinformatics} \\
\textit{Leibniz University Hannover, Germany}\\
0000-0001-5511-9082}
\and
\IEEEauthorblockN{Monika Sester}
\IEEEauthorblockA{
\textit{Institute of Cartography and Geoinformatics } \\
\textit{Leibniz University Hannover, Germany}\\
0000-0002-6656-8809}
}

% \author{%
%   \IEEEauthorblockN{%
%     Yunshuang Yuan\IEEEauthorrefmark{1},
%     Monika Sester\IEEEauthorrefmark{1}
%   }%
%   \IEEEauthorblockA{\IEEEauthorrefmark{1} Institute of Cartography and Geoinformatics, Leibniz University Hannover, Germany}%
% }

\maketitle

\begin{abstract}
Collective Perception has attracted significant attention in recent years due to its advantage for mitigating occlusion and expanding the field-of-view, thereby enhancing reliability, efficiency, and, most crucially, decision-making safety. 
However, developing collective perception models is highly resource demanding due to extensive requirements of processing input data for many agents, usually dozens of images and point clouds for a single frame. This not only slows down the model development process for collective perception but also impedes the utilization of larger models. 
In this paper, we propose an agent-based training framework that handles the deep learning modules and agent data separately to have a cleaner data flow structure. This framework not only provides an API for flexibly prototyping the data processing pipeline and defining the gradient calculation for each agent, but also provides the user interface for interactive training, testing and data visualization. 
Training experiment results of four collective object detection models on the prominent collective perception benchmark OPV2V show that the agent-based training can significantly reduce the GPU memory consumption and training time while retaining inference performance. The framework and model implementations are available at \url{https://github.com/YuanYunshuang/CoSense3D}

\end{abstract}


\begin{figure*}[t]
    \centering
    \includegraphics[width=0.8\linewidth]{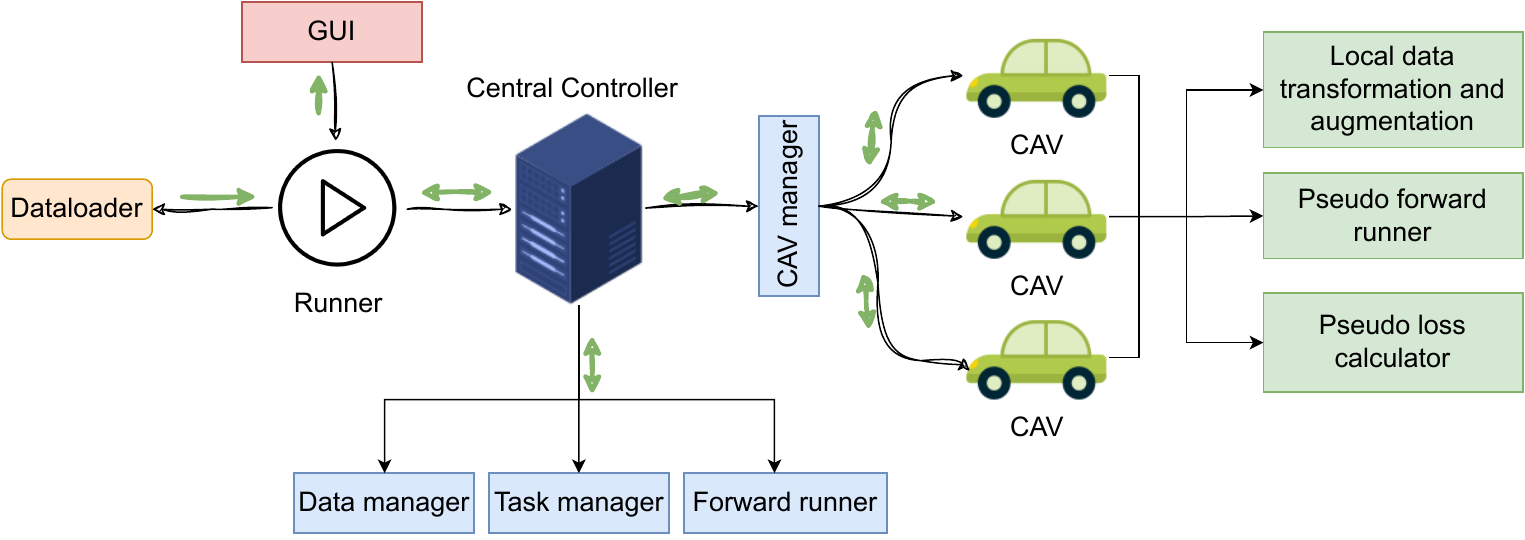}
    \caption{CoSense3D: Agent-based training framework. Black arrows indicate the instruction passing direction, green arrows show the data passing direction.}
    \label{fig:cosense3d_framework}
\end{figure*} 

\section{Introduction}
% collective perception
Reliability and safety are pivotal factors in autonomous driving. Relying solely on the perception capabilities within the view range of a single vehicle is problematic due to potential occlusions. Leveraging collective perception, connected autonomous vehicles (CAV) can mutually share environmental information, enabling observations from multiple vantage points. This approach significantly mitigates occlusions, elevating perception accuracy and enhancing overall safety. 

In addition to safety benefits, autonomous driving offers improved efficiency through optimized trajectory planning. The extended observation range enables autonomous vehicles (AVs) to anticipate and plan more efficiently. However, achieving high accuracy in long-range perception within a single vehicle is challenging because of the limited perception ranges of sensors and increasing sparsity at distant regions. Collective perception markedly enhances accuracy in distant regions by merging denser observation points in these regions from other AVs.

Past studies on collective perception have demonstrated notable success in enhancing the overall accuracy of perception tasks, such as object detection~\cite{fpvrcnn, opencood, v2xvit} and map semantic segmentation in Bird's Eye View (BEV)~\cite{gevbev, cobevt}. Yet, developing deep neural networks for collective perception demands substantial computational power. This requirement involves loading and processing a significantly larger volume of data compared to conventional single-agent-based perception frameworks. To mitigate this challenge and optimize training efficiency while minimizing GPU memory usage, employing more efficient models like EfficientNet~\cite{efficientnet} for image processing and leveraging Sparse 3D convolutions~\cite{spconv, submanifold, minkconv} for point cloud data can be instrumental. Another effective strategy involves utilizing Tensors with lower precision for training and inference~\cite{low-prec}. In addition to these methods, we investigate the impact on perception performance of reducing gradient calculations for specific portions of the input data during training.

To this end, we design a novel framework that optimizes the model development efficiency from two perspectives.
Firstly, existing frameworks for autonomous driving perception such as mmdetection3d~\cite{mmdet3d}  and OpenCOOD \cite{opencood} may encounter a CPU bottleneck and slow down the training process during pre-processing, data transformation and augmentation on a lower-end CPU, particularly evident in collective perception scenarios. In such cases, a single frame of data may contain dozens of images and point clouds, potentially reaching several hundred if the sequential data in temporal dimension is considered. To alleviate this bottleneck without necessitating hardware upgrades, our approach involves migrating the transformation and augmentation processes to the GPU. 
Secondly, computing gradients for all incoming data significantly burdens GPU memory, notably impacting the development of collective perception pipelines, especially with larger models like transformer-based architectures and resource-intensive image processing backbones. To increase development efficiency, we introduce CoSense3D, a specialized framework tailored for collective perception. CoSense3D provides an API for easy definition of data processing pipeline of CAVs, enables flexible control over gradient computation for each CAV, optimizing GPU memory usage and reducing the training time. Regarding each CAV as an agent, we call this agent-oriented framework that defines the behaviour of each agent separately as \textbf{agent-based framework} and the training processing under this framework as \textbf{agent-based learning}.

To assess the effectiveness and performance of our framework in reducing training time and GPU memory usage, we conducted a series of experiments using state-of-the-art models---F-Cooper~\cite{fcooper}, FPVRCNN~\cite{fpvrcnn}, EviBEV~\cite{gevbev} and AttnFusion~\cite{opencood}---for object detection. These experiments are carried out on the collective perception benchmark OpenV2V~\cite{opencood}.

In summary, the contributions of this paper are two fold. Firstly, we proposed and implemented an interactive training, test and visualization framework specially tailored for collective perception of CAVs. Based on this framework, we are able to flexibly define the data processing pipeline and gradient calculation for each CAV. Secondly, based on the above mentioned four models, the results of two groups of comparative experiments with gradient computation for all CAVs versus a limited number of CAVs show that, if the fusion method such as attention fusion with efficient gradient back-propagation is used, reducing the gradient computation for some of the CAVs significantly reduces the GPU memory consumption and training time while maintaining model performance.  

%---------------------------------------------------------------------
\begin{figure*}[t]
    \centering
    \includegraphics[width=0.8\linewidth]{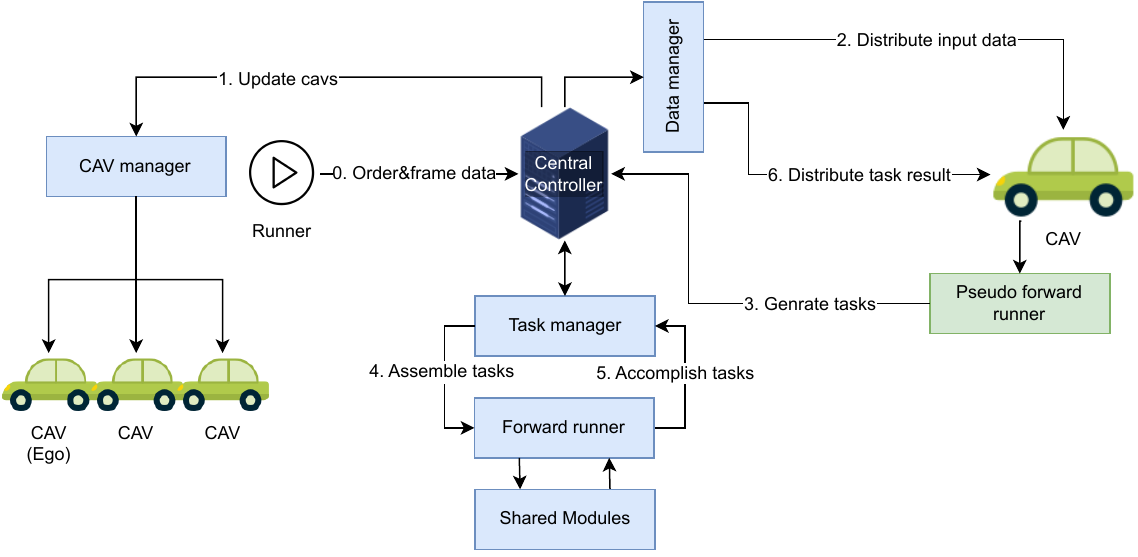}
    \caption{Workflow of CoSense3D Central Controller.}
    \label{fig:central_controller}
\end{figure*}

\section{CoSense3D Framework}\label{sec:framework}
\subsection{Formalization of Collective Perception}	
Consider a set of CAVs denoted as $\textbf{C}={C_1, C_2, ..., C_N}$ engaged in a collective perception scenario. Each CAV is equipped with local sensors. The CAVs $\textbf{C}_\text{nbr}=\{C_j | d(C_i, C_j) < R\}$ in the communication range $R$ of CAV $C_i$ are able to share information to $C_i$, where $d$ is the distance between two vehicles. Assuming $C_i$ is the ego vehicle, at each frame, it first sends its pose $\textbf{P}_i$ as a request to all neighboring vehicles $\textbf{C}_\text{nbr}$, once these vehicles have received the request, they project their local data to the ego vehicle's coordinate system and process these data, which are then sent back to the ego vehicle as a response (CPM). The ego vehicle then fuses all response data from $\textbf{C}_\text{nbr}$ and generates the final object detection result.

\subsection{Framework Structure}
The overall framework is illustrated in figure \ref{fig:cosense3d_framework}. It contains four main modules, namely Dataloader, Graphical user interface (GUI), Runner and Central Controller. The Central Controller is the core module of the CoSense3D framework which contains four sub-modules: CAV manager, Data manager, Task manager and Forward runner. Black arrows indicate the instruction flow, green arrows show the data flow. The framework can run either with or without visualization in the GUI.

\textbf{Dataloader}: The framework standardizes the data loading API for collective perception with a predefined dictionary format to store the meta information in JSON files. With this API, a new dataset can be easily converted to the CoSense3D format without rewriting the PyTorch Dataloader and coping the large media files, such as point clouds and images, to a new data structure. Only the meta information such as scenarios, frames, timestamps, parameters of sensors and the annotations are parsed and saved to CoSense3D format in JSON files.  This standardized Dataloader is able to load images, point cloud data, 2D annotations for images, 3D local annotations for perception without CAV cooperation and 3D global annotations for collective perception.

\begin{figure}
    \centering
     \begin{subfigure}[a]{0.48\textwidth}
         \centering
         \includegraphics[width=\textwidth]{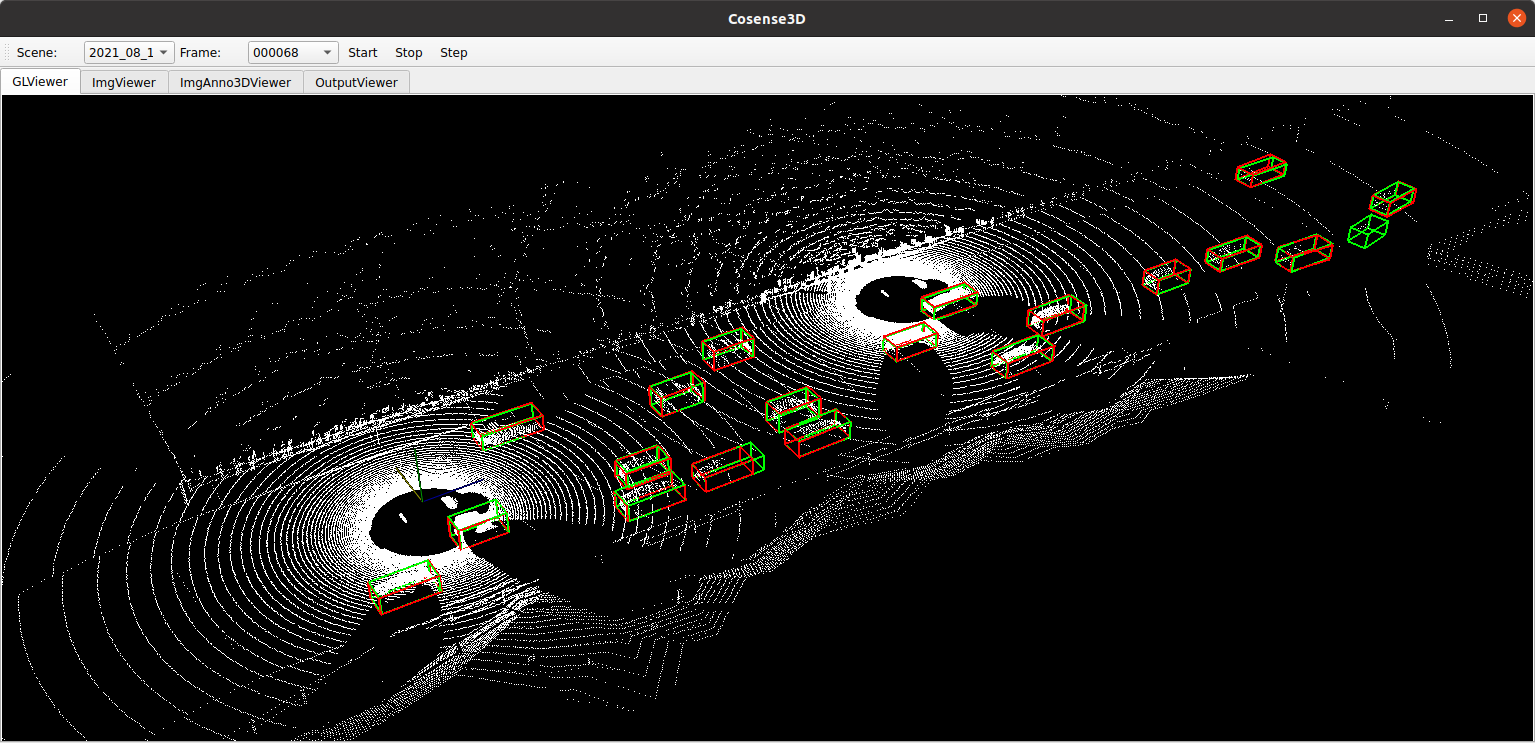}
         \caption{GLViewer}
         \label{fig:glviewer}
     \end{subfigure}
     \hfill
     %  \begin{subfigure}[b]{0.5\textwidth}
     %     \centering
     %     \includegraphics[width=\textwidth]{figs/imgviewer.png}
     %     \caption{ImgViewer}
     %     \label{fig:imgiewer}
     % \end{subfigure}
     % \hfill
     %  \begin{subfigure}[c]{0.5\textwidth}
     %     \centering
     %     \includegraphics[width=\textwidth]{figs/imganno2dviewer.png}
     %     \caption{ImgAnno3DViewer}
     %     \label{fig:imganno3dviewer}
     % \end{subfigure}
     % \hfill
     \begin{subfigure}[d]{0.48\textwidth}
     \centering
     \includegraphics[width=\textwidth]{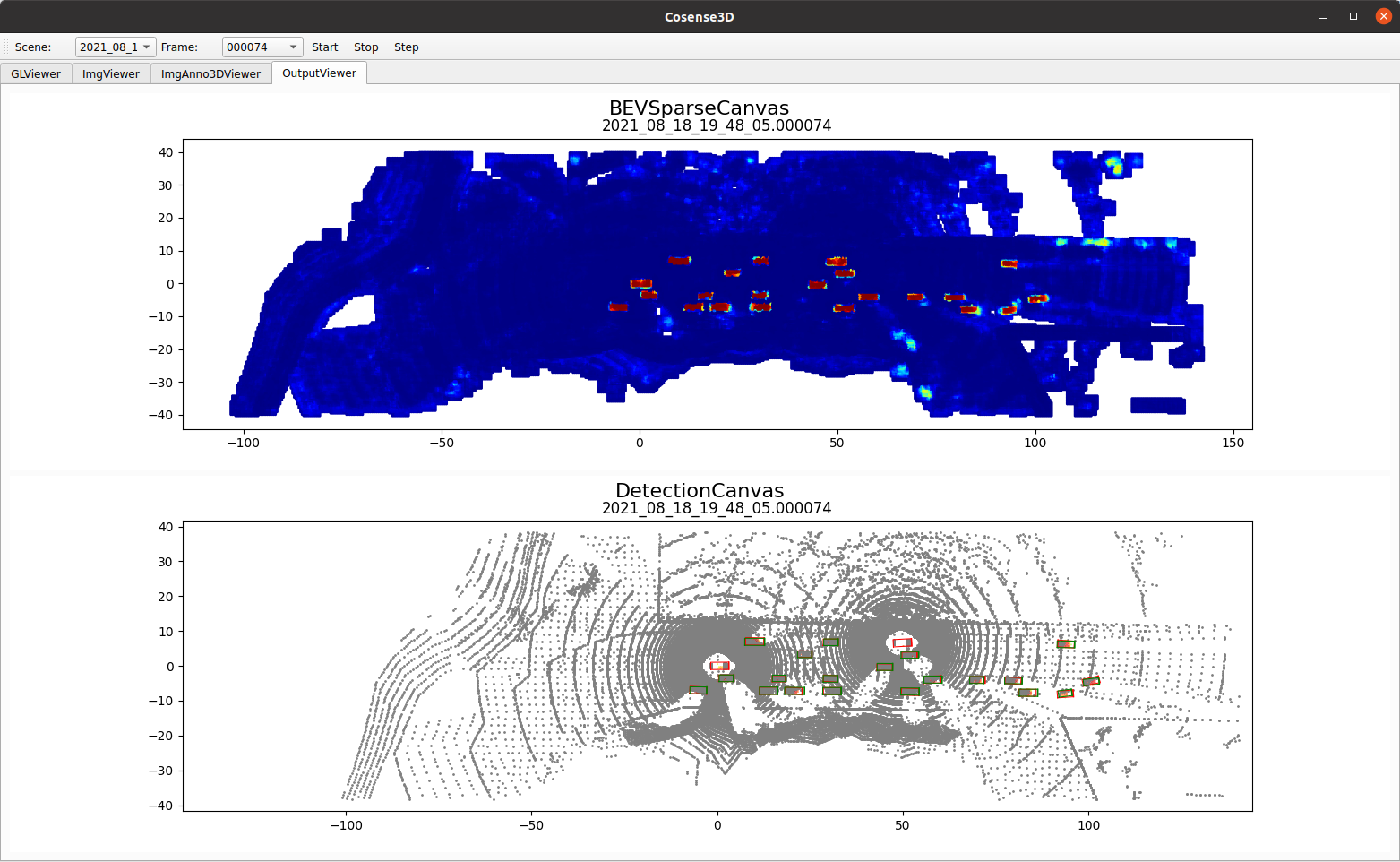}
    \caption{CoSense3D GUI OutputViwer.}
    \label{fig:output_viewer}
     \end{subfigure}
     \hfill
    \caption{CoSense3D GUI.}
    \label{fig:gui}
\end{figure}

\textbf{GUI}: The graphical user interface visualized with the red box in figure \ref{fig:cosense3d_framework} is designed to visualize the training and test data and check the training and test outcomes by one click. This is helpful for loading new datasets and developing new models. Before training on a new dataset, it is necessary to check if the data is converted and loaded correctly. During and after training, visualizing the model output is also helpful to identify the drawbacks and problems of the model and then refine or modify the model accordingly.

As shown in figure \ref{fig:cosense3d_framework}, the GUI can send commands to the runner to start, stop or step the runner process. After each runner step, it updates the visualization modules, 3D GLViewer, ImgViewer, ImgAnno3DViewer and OutputViewer. As shown in figure \ref{fig:glviewer}, GLViewer is a OpenGL-based visualizer for 3D data, annotations (green boxes) and predictions (red boxes). ImgViewer shows image data and the corresponding 2D bounding boxes. ImgAnno3DViewer is used to visualize if the transformations and augmentations of images and 3D annotations are correctly loaded and processed. Each row in ImgViewer and ImgAnno3Dviewer shows the images of a single CAV. After training the model, the OutputViewer can be used to visualize the test result. The OutputViewer can contain multiple canvases which can be customized by the user. Figure \ref{fig:output_viewer} is an example that shows the BEV segmentation (top) and object detection (bottom) result.

\textbf{Runner}: In this framework, three types of Runners are available, namely, TrainRunner, TestRunner and VisRunner. The user can launch these runners with or without GUI.  They are used for training, testing and input data visualization, respectively. Runners manage the frame-wise data and orders dispatching to Central Controller, which then process the orders with the provided frame data accordingly.

\textbf{Central Controller}: As shown in figure \ref{fig:cosense3d_framework}, the Central Controller is the core module of the CoSense3D framework, it communicates with the order-dispatcher (Runner) and the CAVs through its CAV manager. The Data manager is responsible for data gathering and scattering between the central controller and the CAVs. Similarly, the Task manager gathers pseudo tasks generated by CAVs, batches these tasks and dispatches them to the forward runner, which contains all shared deep learning modules for all CAVs, for implementation. In this framework, a standardized CAV prototyping API is provided to allow the user to define the customized workflow for collective perception, including the data augmentations, CAV coordinate transformations, CPM sharing strategies, the forwarding order of the shared neuron network modules and gradient computation strategies of these modules. 

Based on the CAV prototype, the central controller will then implement a standardized pipeline based on the tasks generated by the CAV prototypes as shown in figure \ref{fig:central_controller}. Once the Central Controller receives the order and frame data from the Runner (step 0), the CAV manager will update the CAVs according to the meta information in the frame data and the provided prototype of CAV (step 1). Then the Data manager distributes the input frame data to the updated CAVs (step2). Upon receiving the input data, the CAVs then pre-process the input data and generate tasks and send them back to the Central Controller for processing (step3). To increase the efficiency of the forward process, the Task manager will first summarize the tasks from all CAVs and batch them in two forward steps, one requires gradients, and one without gradient computation, for parallel processing in the Forward Runner (step 4 and 5). After finishing these tasks, the generated results are then distributed back to individual CAVs. 

\section{Experiments}
\begin{figure}[t]
    \centering
    \includegraphics[width=\linewidth]{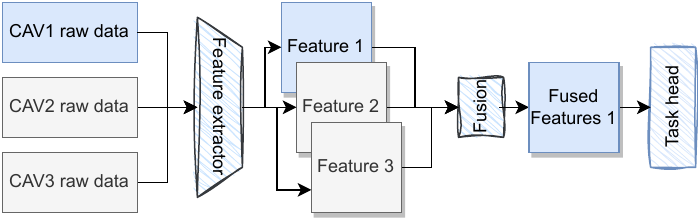}
    \caption{Collective perception pipeline with agent-based training. Blue and gray blocks show the data flow with and without gradient calculation, respectively. Sketched blocks with blues strips are the shared deep learning models.}
    \label{fig:gradient_model}
\end{figure}

To evaluate the efficacy of agent-based training on reducing the GPU memory usage and training time, we conduct two groups of experiments with the state-of-the-art models on OPV2V benchmark. In group one, all models are trained with all gradients enabled. Group two only enables a limited number $\textbf{\small N}_{\textit{grad}}$ of CAVs for gradient calculation. As shown in figure \ref{fig:gradient_model}, we take three CAVs as an example and set $\textbf{\small N}_{\textit{grad}}=1$. Only CAV1 has gradients enabled during training, the features of CAV 2 and 3 are generated with the shared feature extractor with gradient calculation disabled and sent to the CAV 1, where all features are then fused and fed to the task head. 

\subsection{State-of-the-art Networks}
For the above mentioned two groups of experiments, we select the following four representative models with different backbones and fusion methods for cooperative object detection to explore the how these methods influence the detection accuracy and the training efficiency with reduced gradient calculation:

\textbf{AttnFusion}~\cite{opencood} network stands out as one of the top-performing models within the OPV2V benchmark. It uses Voxelnet~\cite{voxelnet} as the backbone network to encode point cloud data, and then convert the sparse Voxels into dense 3D grid to be further encoded with several layers of dense 3D convolutions, which is highly GPU memory demanding. The 3D voxel features are then concatenated to obtain the final BEV features map with shape $(H, W, d)$ and shared to the ego vehicle, where $H$ and $W$ represent the height and width of the feature map, and $d$ signifies its depth. Subsequently, this feature map is shared with the ego vehicle. To fuse these feature maps, as implied by its name, AttnFusion incorporates an attention module which consolidates the BEV feature maps from all $M$ neighboring CAVs, generating a 4D Tensor with shape $(M, H, W, d)$. The attention mechanism functions by attending to and blending features along the first axis among the $M$ CAVs. At last, it uses an anchor-based detection head to generate the bounding box classification and regression result.

\textbf{F-Cooper} is the earliest model applied to feature-level cooperative perception. It first encode the voxels into 2D BEV feature maps, and then fuse these maps by a Maxout operation, which takes the maximum feature values among all feature maps coming from CAVs as the fused features. Afterwards, an anchor-based detection head is used to predict the objects in the scenario.

\textbf{FPVRCNN} is a two-stage detector that requires the least communication bandwidth with adequate performance within the OPV2V benchmark. As the name indicates, this framework is built based on PVRCNN~\cite{pvrcnn}. However, in order to get better proposal detections for sharing, it uses a spatial-semantic feature aggregation (SSFA) module proposed in CIASSD~\cite{ciassd} to enhance the BEV feature encoding for the first-stage detection. Based on these detections, FPVRCNN samples some keypoints within the detected bounding boxes, and aggregate deep features of different resolutions for each keypoints using the Voxel Set Abstraction (VSA) Module of PVRCNN, the keypoints features along the  proposal bounding boxes are then shared and fused in the second stage to obtain refined bounding boxes.

\textbf{EviBEV} is a fully sparse and highly efficient framework~\cite{gevbev} for LiDAR-based cooperative BEV semantic segmentation and 3D object detection. It encodes the points with a Minkowski-convolution-based Unet. To enhance the connectivity between the sparse voxels during the convolutions, it additionally augments the original point clouds with free space points along the LiDAR casting ray before feeding the points to the Unet. In addition, several layers of coordinate-expandable sparse convolutions a utilized after the Unet to compress the features along the vertical axis and dilate the sparse coordinates so that they can have a better coverage over the centers of visible object and then perform the final detection. Different from the networks mentioned above, this work uses the anchor-free center-point~\cite{centerpoint} head for object detection over the sparse BEV coordinates. 

\subsection{Dataset}
In this study, we utilize the simulated dataset OpenV2V~\cite{opencood} to assess the efficiency of our framework. OpenV2V serves as a prominent benchmark specifically curated for collective perception tasks. It comprises 44 scenes and 6765 training frames, along with 16 scenes and 2170 frames designated for testing purposes. These scenes encompass diverse driving scenarios, spanning urban environments, rural areas, and highways across nine simulated cities.

Adhering to the official settings, our parameters are configured as follows: the communication range $R$ is set to $70$ meters, the maximum number of cooperative vehicles $\textbf{C}_\text{nbr}$ is set to 7, and the detection ranges span $[-140.8, 140.8]m$, $[-38.4, 38.4]m$, and $[-3.0, 1.0]$ along the x-, y-, and z-axes, respectively.

\subsection{Experiment Settings}
All models are trained on the a single Nvidia RTX3090Ti GPU and an Intel Core i7-8700 CPU for 50 epochs. Batch size is set to 2. All models use Adam optimizer with starting learning rate of $0.002$ and weight decay of $1e-4$. The learning rates reduce at epoch $15$ and $30$ with the multiplication factor of $0.1$. During training, the input data are augmented with rotation along z-axis with a random angle in the range of $[-90, 90]^{\circ}$, random flipping along x- and y-axis, and scaling with a ratio in the range of $[0.95, 1.05]$. Additionally, the ground truth bounding boxes with less than two points are removed during training process.

\section{Result and Evaluation}

\begin{table}[t]
\centering
 {\fontsize{8}{9.6}\selectfont
\begin{tabular}{c|c|c|c|c|c}
\hline

 \rule{0pt}{2ex}\textbf{\small Model} & $\textbf{N}_{\textit{grad}}$ & AP$@0.7$ & AP$@0.5$ & AP$^{-}$$@0.7$ & AP$^{-}$$@0.5$ \\
\hline
\multirow{2}{*}{\small F-Cooper}
 & all & \textbf{82.2} & \textbf{89.9}  & \multirow{2}{*}{-13.5} &\multirow{2}{*}{-6.4}\\ 
 & 2 & 68.7 & 83.5 &&\\
 \hline
\multirow{2}{*}{\small FPVRCNN}
 & all & 84.0 & 87.3 & \multirow{2}{*}{\textbf{+0.9}} &\multirow{2}{*}{\textbf{+0.5}}\\
 & 1 & \textbf{84.9} & \textbf{87.8} &&\\
 \hline
\multirow{2}{*}{\small EviBEV}
 & all & \textbf{84.1} & \textbf{91.1} & \multirow{2}{*}{-4.6} &\multirow{2}{*}{-2.0}\\
 & 1 & 79.5 & 89.1  &&\\
 \hline
\multirow{2}{*}{\small AttnFusion}
 & all & \textbf{87.6} & 92.3 & \multirow{2}{*}{-0.5} &\multirow{2}{*}{\textbf{+0.3}}\\
 & 1 & 87.1 & \textbf{92.6}  &&\\
\hline
\end{tabular}}
\caption{Average precision of object detection on OPV2V benchmark with different gradients configurations. AP$@iou$: average precision at IoU threshold $iou$. AP$^{-}$$@iou$: the AP drop of reduced gradient computation in comparison to the full gradient computation.}
\label{tab:agent_based}
\end{table}

\subsection{Average Precision (AP) of Object detection}\label{sec:eval1}
The AP of object detection of the selected models are listed in Table \ref{tab:agent_based}. Two commonly used Intersection over Union (IoU) thresholds are used for calculating the APs, namely $0.5$ and $0.7$. They are notated with AP$@0.5$ and AP$@0.7$, respectively. AP$^{-}$ in the last two columns of table \ref{tab:agent_based} means the performance drop of reduced gradient computation in comparison to the full gradient computation.
$\textbf{\small N}_{\textit{grad}}$ is the number of CAVs that require gradient calculation during training. $\textbf{\small N}_{\textit{grad}}=\text{all}$ means all CAVs require gradients. Different from the other three models, we set $\textbf{\small N}_{\textit{grad}}$ of F-Cooper to $2$ in the second group of experiments. This adjustment was made as $\textbf{\small N}_{\textit{grad}}=1$ in this case exhibited notably poor performance. In general, F-Cooper performs worse than the other models and is very sensitive to the number of gradient-enabled CAVs. This is caused by the Maxout operation, which filters out some of the BEV features coming from a gradient-enabled CAV and keeps the non-gradient features. In this way, the downstream modules after Maxout are not able to back-propagate all gradients back to the upstream modules that encode point clouds into BEV features. 

In comparison, EviBEV has significantly smaller performance drops than F-Cooper when using ego-agent-based training.  However,  its performance is worse than FPVRCNN and AttnFusion. This is because it uses naive fusion to merge the sparse BEV features from all CAVs,  which is similar to Maxout that requires no gradient calculation. Specifically, it involves mixing all BEV feature points from all CAVs and randomly sampling one feature point at a particular BEV position in cases where multiple CAVs have overlapping feature points at that BEV position. This operation also weakens the back-propagation for upstream modules because some of the gradient-enabled features are dropped.

Differently, both FPVRCNN and AttnFusion use gradient-enabled modules for fusing the features coming from all CAVs. FPVRCNN merges all keypoint features with a RoI grid pooling layer while AttnFusion fuses BEV maps from different CAVs by an attention module that attends the features along all CAVs. The performance drops of these two models with only the gradient calculation for ego agent are not noticeable. In some cases, the performance is even enhanced.  For example, the APs at the IoU threshold of $0.5$ for FPVRCNN and AttnFusion with ego-agent-based training both surpasses that with all-agent-based training by a small margin. 

From the experiments result shown in table \ref{tab:agent_based}, we conclude that agent-based training may deteriorate the detection performance if the model uses a non-learnable module that partially drops the upstream features from gradient-enabled CAVs. However, if a more effective learnable fusion module is used to merge all upstream features without dropping gradient-enabled features, we are able to use the more efficient agent-based training framework to accelerate the training and model development process without noticeable performance drop.

\subsection{Efficiency of Agent-based Learning}

\begin{figure}[t]
    \centering
    \includegraphics[trim=0 0 0 7,clip, width=1.0\linewidth]{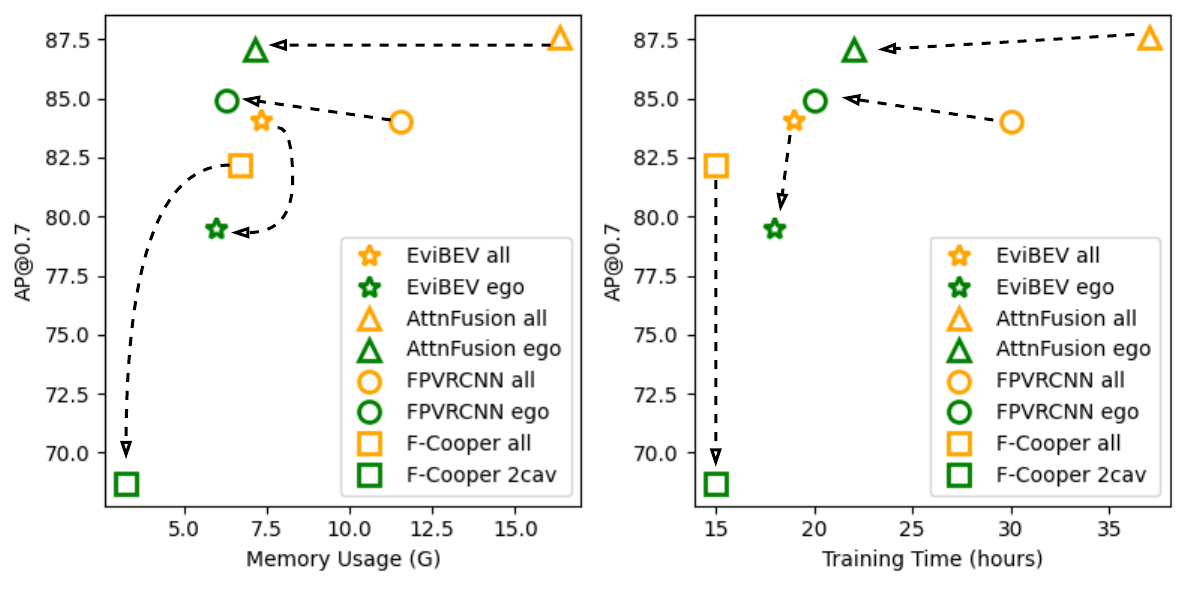}
    \caption{Correlation between object detection performance and GPU memory usage (left) as well as training time (right) with different gradient configurations.}
    \label{fig:mem_time}
\end{figure}
To quantitatively evaluate the efficiency of our proposed agent-based training framework, we illustrate the detection AP versus the memory usage and training time as shown in figure \ref{fig:mem_time}. Yellow markers show the results of training with all gradients enabled, and green markers the results of agent-based training. The results of different models are illustrated with different shape of markers. It is readily evident that agent-based training can significantly reduce the GPU memory usage and the training time, especially for larger models. For example, AttnFusion consumes about 16.7$G$ GPU memory with gradient enabled for all CAVs, while agent-based training only uses 7.3$G$, and thus reduces about 56\% of GPU memory usage. 

The reduced load of gradient calculation directly impacts the training time. This can be identified by the similar patterns of the yellow and the green markers between the left and the right plot in figure \ref{fig:mem_time}. The reduction ratio in GPU memory usage closely mirrors that of the training time except F-Cooper. The memory usage of F-Cooper is nearly halved, whereas the training time keeps nearly unchanged. This result testifies again the conclusion drawn from the failure case of F-Cooper discussed in section \ref{sec:eval1} which is caused by Maxout operation. The memory usage is reduced because the gradients for non-learnable CAVs are not cached during the forward run. However, for all configurations of $\textbf{\small N}_{\textit{grad}}$, F-Cooper only back-propagates $m \leq h\cdot w\cdot d$ gradients to the upstream modules before Maxout, where $m$ is the number features with gradients back-propagated and $h$, $w$, $d$ is the height, width and depth of the BEV feature map of a single CAV, respectively. Increasing $\textbf{\small N}_{\textit{grad}}$ for F-Cooper only makes $m$ to be closer to the maximum value $h\cdot w\cdot d$. In contrast, gradients of learnable fusion method such as attention fusion used by AttnFusion are fully propagated back to the learnable CAVs. Mathematically, $m=\textbf{\small N}_{\textit{grad}}\cdot h\cdot w\cdot d$ gradients are back-propagated, where $m$ is linearly related to the number of gradient-enabled CAVs. Therefore, decreasing $\textbf{\small N}_{\textit{grad}}$ of AttnFusion can linearly reduce the gradients calculation hence reducing the training time. This effect can hardly be observed by F-Cooper because it only reduces a very small portion of gradients of a single BEV feature map. Similarly, EviBEV only reduces the training time by less than one hour because of the random sampling of the naive fusion module. In addition, the memory reduction of EviBEV by agent-based learning is noticeable less than the other three models because it is a fully-sparse model that is already very efficient with respect to the increasing number of CAVs.

Therefore, agent-based training can significantly reduce the memory usage and training time without noticeable performance drops if an appropriate fusion method without dropping learnable features is used. In this case, the memory usage and training time of dense models is roughly positively proportional to the number of the learnable CAV $\textbf{\small N}_{\textit{grad}}$. However, for sparse models, this relationship is non-linear and related to the actual size of the sparse tensors.

\section{Conclusions}
We proposed an efficient agent-based training framework that is specifically designed for multi-agent collective perception.
It aims to reduce the GPU memory usage and the training time for the development of collective perception models, which usually needs to process enormous image and point cloud data for multiple CAVs. This framework explicitly separates the data management of different CAVs and deep learning modules to provide a clean workflow to flexibly define the data processing pipeline and adjust the gradient configuration for each CAV. Two groups of comparative experiments with full gradient calculation versus reduced number of CAVs for gradient calculation have shown the training efficiency and performance with our proposed framework. The result shows that reducing the number of CAVs for gradient calculation can significantly save GPU memory and training time without noticeable performance drop, if an appropriate fusion module without dropping learnable features is used. On the OPV2V dataset, the training resources needed has been reduced by more than a half for the best-performing model AttnFusion.
In addition, we also showed two failure cases, F-Cooper with Maxout fusion and EviBEV with Naive mix-up fusion, and discussed the reason behind. Similar operations can be avoided by model design if our proposed agent-based training is used.

%%%%%%%%%%%%%%%%%%%%%%%%%%%%%%%%%%%%%%%%%%%%%%%%%%%%%%%%%%%%%%%%%%%%%%%%%%%%%%%%
\section*{Acknowledgment}
This work is supported by the projects \href{www.socialcars.org}{DFG RTC1931 SocialCars}.

\bibliographystyle{IEEEtran}
\bibliography{IEEEexample}

\end{document}